%% file: acl_latex.tex
\pdfoutput=1

\documentclass[11pt]{article}

\usepackage[]{emnlp2022}

\usepackage{times}
\usepackage{latexsym}

\usepackage[T1]{fontenc}

\usepackage[utf8]{inputenc}

\usepackage{microtype}
\usepackage{adjustbox}
\usepackage{booktabs}
\usepackage{multirow}
\usepackage{amsmath}
\usepackage{amssymb}
\usepackage{amsthm}
\usepackage{enumitem}
\usepackage{bm}
\usepackage{bbm}
\usepackage{algorithm}
\usepackage{algcompatible}
\usepackage{algpseudocode}
\usepackage{xcolor}
\usepackage{xspace}

\def\figref#1{Fig.~\ref{#1}}
\def\secref#1{Sec.~\ref{#1}}
\def\appendixref#1{Appendix~\ref{#1}}
\def\tabref#1{Table~\ref{#1}}
\def\eqnref#1{Eqn.~\ref{#1}}
\def\algref#1{Alg.~\ref{#1}}
\def\condref#1{Hypothesis~\ref{#1}}
\def\propertyref#1{~\ref{#1}}

\newcommand{\met}{\textbf{M}utual \textbf{E}xclusivity \textbf{T}raining (MET)\xspace}
\newcommand{\abbrmet}{MET\xspace}

\newcommand{\addjump}{\textit{Jump}\xspace}
\newcommand{\adddax}{\textit{Dax}\xspace}
\newcommand{\aroundright}{\textit{Around Right}\xspace}

\newlist{Properties}{enumerate}{2}
\setlist[Properties]{label=Property \arabic*, font=\textbf, itemindent=*, align=left}

\newlist{Definitions}{enumerate}{2}
\setlist[Definitions]{label=Definition \arabic*, font=\textbf, itemindent=*, align=left}

\newlist{Conditions}{enumerate}{2}
\setlist[Conditions]{label=Hypothesis \arabic*, font=\textbf, itemindent=*, align=left}

\newlist{Conditions2}{enumerate}{2}
\setlist[Conditions2]{label=2.\Alph*, font=\textbf, itemindent=*, align=left}

\newtheorem{definition}{Definition}

\newtheorem{property}{Property}

\newtheorem{condition}{Hypothesis}

\title{Mutual Exclusivity Training and Primitive Augmentation \\to Induce Compositionality}

\author{Yichen Jiang\thanks{\:\:Equal contribution.} \ \ \ \ \ \ \ \ Xiang Zhou\footnotemark[1] \ \ \ \ \ \ \ \ Mohit Bansal \\
UNC Chapel Hill \\
  \texttt{\{yichenj, xzh, mbansal\}@cs.unc.edu} \\
}

\begin{document}
\maketitle
\input{main_tex/0abstract}
\input{main_tex/1intro}

\input{main_tex/2background}

\input{main_tex/3methods}

\input{main_tex/4experiments}
\input{main_tex/5related}
\input{main_tex/6conclusion}
\input{main_tex/7limitations}

\input{main_tex/8ethics}

\section*{Acknowledgements}
We thank the reviewers for their helpful comments. This work was supported by ONR Grant N00014-18-1-2871, NSF-CAREER Award 1846185, and DARPA MCS Grant N66001-19-2-4031.
The views are those of the authors and not of the funding agency.

\bibliography{anthology,custom}
\bibliographystyle{acl_natbib}

\appendix

\section*{Appendix}
\label{sec:appendix}
\input{appendix_tex/1methods}
\input{appendix_tex/2experiments}

\input{appendix_tex/3ulcomparison}
\end{document}

%% file: main_tex/0abstract.tex
\begin{abstract}
Recent datasets expose the lack of the systematic generalization ability in standard sequence-to-sequence models. 
In this work, we analyze this behavior of seq2seq models and identify two contributing factors:
a lack of mutual exclusivity bias 
(i.e., a source sequence already mapped to a target sequence is less likely to be mapped to other target sequences),
and the tendency to memorize whole examples rather than separating structures from contents. 
We propose two techniques to address these two issues respectively: \textit{Mutual Exclusivity Training} that prevents the model from producing seen generations when facing novel, unseen examples via an unlikelihood-based loss;
and \textit{prim2primX data augmentation} that automatically diversifies the arguments of every syntactic function to prevent memorizing and provide a compositional inductive bias without exposing test-set data.
Combining these two techniques, we show substantial empirical improvements using standard sequence-to-sequence models (LSTMs and Transformers) on two widely-used compositionality datasets: SCAN and COGS.
Finally, we provide analysis characterizing the improvements as well as the remaining challenges, and provide detailed ablations of our method.\footnote{The code for our EMNLP 2022 paper is available at \url{https://github.com/owenzx/met-primaug}}

\end{abstract}

%% file: main_tex/1intro.tex
\section{Introduction}
Human intelligence demonstrates \textit{systematic compositionality}, the algebraic capacity to understand and produce a potentially infinite number of novel combinations of known components~\cite{chomsky1957syntactic,montague1970universal}, in comprehending and generating natural language.
However, a number of recent datasets, e.g., SCAN~\cite{lake2018generalization}, COGS~\cite{kim-linzen-2020-cogs}, etc., provide clear evidence of the lack of systematic compositionality in state-of-the-art neural networks.
By analogy with what humans do in meaningful learning~\cite{ausubel1963psychology, shi2020scan}, methods to solve this problem can be categorized into \textit{deductive learning} and \textit{inductive learning} methods. 
In \textit{deductive learning}, the learner is first introduced to a general rule and then followed by specific examples where the rule is applied~\cite{thornbury1999teach}, while in \textit{inductive learning}, the learner is first presented with abundant examples so that the learner can automatically infer the general rule by itself.
Under the context of compositional generalization, \textit{deductive learning} approaches directly inject useful priors~\cite{li-etal-2019-compositional,russin-etal-2020-compositional,liu2020compositional,liu-etal-2021-learning-algebraic} to the model;
while \textit{inductive learning} approaches utilize data augmentation\cite{andreas-2020-good,akyurek2021learning} to provide more accurate examples that facilitate the learning of compositionality.
In this work, we focus on designing methods that do not require any specific architectural changes, so it is applicable to standard sequence-to-sequence (seq2seq) models, and propose both deductive and inductive learning paradigms. 

Our deductive method is derived from an observation that when facing compositionally novel examples, models have a tendency to generate old patterns seen during training, hence leading to numerous mistakes. 
Despite being a major challenge for models, humans avoid these mistakes by exploiting the \textit{mutual exclusivity (ME) bias}~\cite{markman1988children}, i.e., if a concept is already associated with one expression, humans are less likely to associate a new expression to that concept.
Consequently, when given an unseen test input (expression), humans have the prior that this new input should not be mapped to a seen output (concept) that is already mapped to another input expression.
Inspired by the ME bias, we propose a novel training framework \met. 
Given an input-output pair, \abbrmet encourages the model to assign a high probability to the output given the input, and to \textit{not} generate this output given any other inputs. 
We achieve this by first creating a randomly perturbed version of the input, and then adopting the unlikelihood loss~\cite{welleck2019neural} on the perturbed input to penalize the generation of the original output.
Finally, we train the model using a joint loss of MLE on the original batch and the unlikelihood loss.
Additionally, we explore another variant \abbrmet-Meta, where we embed the unlikelihood loss as the meta generalization objective of the MAML framework (visualized in~\figref{fig:meta-unlike}). 
This meta-learning objective will provide more regularization on the gradient updates instead of the final optimization target.
On COGS and SCAN, we see substantial improvements from both variants of \abbrmet as long as the baseline seq2seq model has acceptable performance. \abbrmet improves a Transformer model from 76.1\% to 80.6\% on COGS, and improves an LSTM model from 13.5\% to 38.7\% on SCAN MCD2. 
The two \abbrmet variants also show interesting behavior differences. 
For Transformers, \abbrmet shows advantages on COGS, and \abbrmet-Meta show advantages on most of the SCAN tasks.

\begin{table*}[t]
\centering
\begin{scriptsize}
\begin{tabular}[t]{c|c|c|c}
\toprule
 \centering \multirow{3}{*}{\centering \small Function $f$} & \multicolumn{3}{c}{\small Examples} \\
 \cmidrule{2-4}
  & \textbf{Input} & \textbf{Program} & \textbf{Output} \\
\midrule
 $\mathrm{Rev}(x_1, x_2) = x_2 + x_1$ & walk left & $\mathrm{Rev} (\mathrm{walk}, \mathrm{left})$ & TL WALK \\   
 $\mathrm{Around}(x_1, x_2) = (x_2 + x_1 ) * 4$ & walk \textbf{around} left & $\mathrm{Around} (\mathrm{walk}, \mathrm{left})$ & TL WALK TL WALK TL WALK TL WALK  \\ 
$\mathrm{Twice}(x) = x * 2$ & walk opposite left \textbf{twice} & $\mathrm{Twice}(\mathrm{Oppo}(\mathrm{walk}))$ & TL TL WALK TL TL WALK \\
\bottomrule
\end{tabular}
\end{scriptsize}
\vspace{-6pt}
\caption{Syntactic function examples in SCAN~\cite{lake2018generalization}. ``$+$'' means concatenation. ``$x * 2$'' means replicating $x$ twice. A full table containing all the syntactic functions in SCAN is in the Appendix.
}
\vspace{-9pt}
\label{table:scan_functions}
\end{table*}

Despite the above improvements, a Transformer model equipped with \abbrmet may still struggle with very poor baselines. 
We hypothesize that this phenomenon may partially be because that the syntactic function to be generalized has not been applied to a sufficient number of distinct arguments in the training set.
For instance, in the SCAN \addjump, the function ``\textit{around left}'' is only paired with four different lexical arguments (``\textit{walk, run, look, turn}''). 
This lack of argument diversity makes it easier for the model to \textit{memorize} all these patterns, but harder to infer the syntactic rules with \textit{deductive learning}.
To alleviate this memorization problem, we propose \textbf{prim2primX}, a data augmentation method to promote generalization by automatically generating a large set of new lexical arguments for each syntactic function. 
As the number of distinct arguments for each function increases, memorizing
each single example independently becomes more challenging, 
while the difficulty of understanding the compositional structure remains the same, which makes it a more encouraging behavior.
Specifically, the prim2primX procedure first builds a lexicon using a dataset-agnostic, rule-based word alignment algorithm.
This lexicon contains the mapping between the input and the output form of each argument (e.g., ``\textit{run}$\mapsto$\texttt{RUN}''), 
while ignoring functional words (e.g., ``\textit{around}'') that only decide the syntactic structure of the outputs.
Then, to enrich the set of lexicons, we mutate each primitive (``\textit{run}$\mapsto$\textit{run0}; \texttt{RUN}$\mapsto$\texttt{RUN0}''),
and create new examples by swapping the primitive argument (prim) with its mutated form (primX) on both sides.
Our experiments (see \secref{sec:exp}) provide evidence that supports our claim above:
with 15 new primitives (5 per original primitive) added to the SCAN lexicon, the accuracy on the SCAN \addjump test set reaches 63.65\% from the original performance of 3.49\%.
On COGS, with 2 new primitives per original one, prim2primX
also improves the Transformer performance from 76.14\% to 80.07\% accuracy.
Furthermore, we show that combining prim2primX and \abbrmet 
can achieve further improvement. Our prim2primX+\abbrmet achieves 81.1\% accuracy on COGS, and prim2primX+\abbrmet-Meta achieves 74.0\% on SCAN \addjump.
Finally, we also provide detailed ablations about each component in our proposed methods, and an analysis characterizing our improvements and remaining challenges.

Overall, our contribution is two-fold: 
(1) we propose \abbrmet, a novel, deductive training framework to inject mutual exclusivity bias into models; 
and (2) we propose prim2primX, a data augmentation method that automatically creates new arguments to facilitate inductive compositionality learning. 
Both methods and their combination show substantial empirical improvements on SCAN and COGS. 
Moreover, our methods are data and model-agnostic, and do not leak any test examples, so the improvements reveal the compositional generalization potential of general seq2seq models.

%% file: main_tex/2background.tex
\section{Background}

\subsection{Compositional Generalization Challenge}
\label{ssec:comp_gen_challenge}

Compositional generalization challenges test models with \textit{unseen combinations of seen structures and seen contents}.
While neural models perform well on in-distribution examples, they fail on most compositionality challenges with a distributional shift between training and test examples even when some surface statistics (e.g., word frequency) are similar. 
One example dataset used in this work is \textbf{SCAN}~\cite{lake2018generalization}, which includes paired natural language commands and action sequence outputs.
We can represent this task as applying \textit{syntactic functions} to the arguments.
\begin{definition} \label{def:function}
A \textbf{\textit{syntactic function}} $f$ is a symbolic function which maps certain input patterns to corresponding output structures.
\end{definition}
\tabref{table:scan_functions} lists some example syntactic functions and corresponding input-output pairs in SCAN.
The main challenge in SCAN \addjump is to apply functions like $\mathrm{Around}(x_1, \mathit{left})$ to a lexical argument ``\textit{jump}'', while such combination does not appear in the training set.
Another dataset we use in this work is \textbf{COGS}~\cite{kim-linzen-2020-cogs}, which
requires parsing a diverse set of natural language sentences into their corresponding logical forms.
COGS 
raises five different generalization challenges:
syntactic functions with novel (1) primitives or (2) modified phrases;
(3) deeper recursion; 
(4) alternative verb argument; and 
(5) novel identification of a verb. 
Challenges 1, 4, and 5 require generalizing to unseen lexicons, 
while challenges 2 and 3 require generalizing to unseen structures.

\subsection{Limitations of Standard Seq2Seq Models} 
\label{sec:error_ana}
Standard seq2seq models often struggle facing compositional challenges
and fall into a common mistake pattern.
For example, given a test command ``\textit{jump left twice}'' that requires generalizing a syntactic function to ``\textit{jump}'',
Transformer sometimes generates the correct output structure but mistakenly picks another familiar primitive (e.g., ``walk'')
over ``\textit{jump}'', yielding ``\texttt{TL \textbf{WALK} TL \textbf{WALK}}''.
This suggests that during the training, Transformer is not separately mapping the function ``$x$ \textit{left twice}'' to the output structure ``\texttt{TL} $x$ \texttt{TL} $x$''; and the argument ``\textit{walk}'' to the action ``\texttt{WALK}''.
As a result, when facing a novel command (e.g., ``\textit{jump left twice}''), in 79.8\% of test cases, the model does not generate ``\texttt{JUMP}'' but instead repeats a seen training-set output.
Additionally, in more than 63\% of the test examples, the model cannot even correctly interpret the functional structure, e.g., it generates the structure for ``\textit{jump around left and run}'' when it is required to ``\textit{jump around left and run thrice}''.
In contrast, humans 
(1) have a natural prior to not map two different sentences to the same point in the output space~\cite{markman1988children}; and 
(2) are able to correctly separate syntactic structures from the actual contents~\cite{DBLP:conf/cogsci/LakeLB19}.
In the next section, we will show how we relieve this problem.

%% file: main_tex/3methods.tex
\section{Methods}

\subsection{Injecting Mutual Exclusivity Bias}
\label{ssec:unlikelihood_training}
One crucial factor for humans to prevent the mistakes in \secref{sec:error_ana} is the \textit{mutual exclusivity bias}~\cite{markman1988children}: if a concept is already associated with one expression, a human is less likely to associate a new expression to that concept. 
Under the context of seq2seq tasks, this bias should lead to the following behavior: in a bijective function task (i.e., the input and the output have one-to-one correspondence), when facing an unseen input, the model should not produce a seen output that is already associated with a seen input.
Formally, we can represent it as: 
\vspace{-1pt}
\begin{property}
\label{prop:mutual_exclusive} Given a bijective training set $D$ containing $n$ input-output pairs $(x_i, y_i)$. For a given test example $\hat{x}$, if $\forall i, \hat{x}\neq x_i$, then the label for the test example $\hat{y}$ will have the property $\forall i, \hat{y}\neq y_i$.
\end{property}
\vspace{-1pt}
Following this formalization, we propose a novel framework \met to inject this property into standard seq2seq models.

\paragraph{\abbrmet with Unlikelihood Loss.}
We notice that Property\propertyref{prop:mutual_exclusive} can be achieved in a model-agnostic way by penalizing the predicted likelihood on $\hat{y}$, hence becoming a variant of \textit{unlikelihood training}~\cite{welleck2019neural}.  
Intuitively, during training, for a given mini-batch $(x, y)$, we sample some negative inputs $\tilde{x}\neq x$, which are similar to the original input $x$ but \textbf{do not} map to the corresponding target $y$; and then we encourage the model to \textbf{not} generate $y$ given $\tilde{x}$. 
Specifically, we train our model on the sum of maximum likelihood estimation (MLE) and unlikelihood loss:
\vspace{-5pt}
\begin{equation*}
\label{eq:ul}
    \mathcal{L}_{\mathrm{MLE}} = - \log (p(y|x)), 
    \mathcal{L}_{\mathrm{UL}} = - \log (1-p(y|\tilde{x}))
\vspace{-5pt}
\end{equation*}
Notably, this unlikelihood loss $\mathcal{L}_{\mathrm{UL}}$ is different from the loss used in \citet{welleck2019neural}, as the original unlikelihood is an average over all possible words, while the unlikelihood loss in Eq.~\ref{eq:ul} is operated at the sentence level. We adopt this change since sentence-level unlikelihood provides a suitable amount of regularization in our setting, while the regularization from the original word-level loss is too strong.
See Appendix~\ref{app:met} for more discussions, proof-of-concept experiments and ablation results.

\begin{figure}[t!]
\begin{center} 
\includegraphics[width=0.4\textwidth]{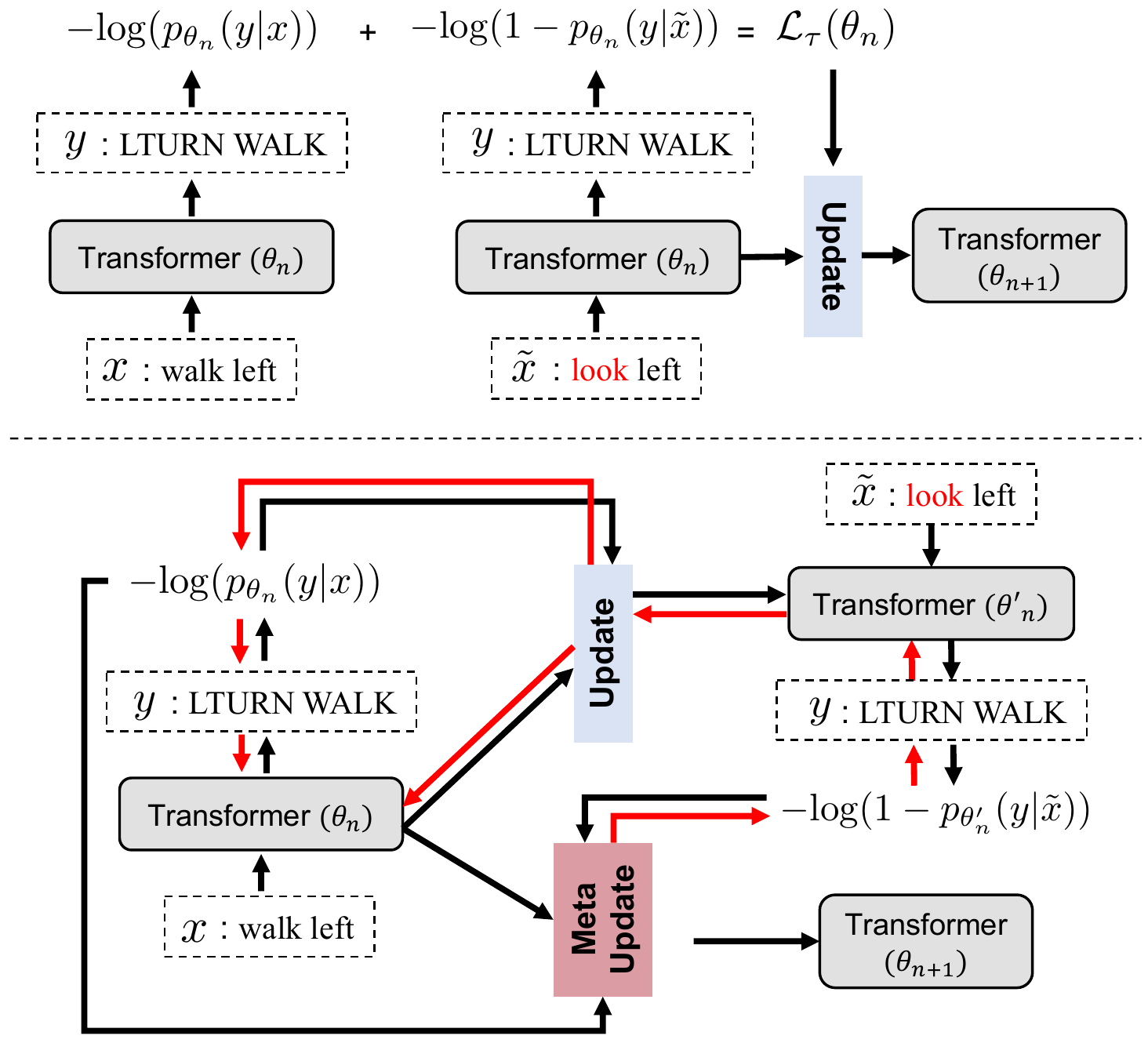}
\end{center} 
\vskip -0.1in
\caption{\textbf{Upper}: \abbrmet.
\textbf{Lower}: \abbrmet-Meta first updates the model with the likelihood loss, then performs a meta update with the sum of likelihood loss and meta-unlikelihood loss.
The \textcolor{red}{red} arrows track the gradients generated by the meta-unlikelihood loss.  
\label{fig:meta-unlike}
}
\vspace{-10pt}
\end{figure}

\paragraph{\abbrmet-Meta.} \abbrmet encourages Property\propertyref{prop:mutual_exclusive} in \textit{the final trained model}. 
Alternatively, Property\propertyref{prop:mutual_exclusive} can also be viewed as a regularization on the \textit{learning behavior}, i.e., once a model is trained on a pair $(x, y)$, it will give a lower probability to the pair with the same output but a different input $(\tilde{x}, y)$.
We propose a variant: \abbrmet-Meta that achieves this effect by embedding the unlikelihood loss in a MAML~\cite{finn2017model,conklin-etal-2021-meta} framework.
In MAML, each batch consists of two sets of examples, `meta-train' and `meta-test'.
MAML encourages the model optimized on the meta-train examples to also optimize the loss on the meta-test, thus encouraging more generalizable behaviors. 
In \abbrmet-Meta, we use unlikelihood loss as the loss on meta-test.
Specifically, with a sampled mini-batch $(x, y)$ (meta-train), we first calculate the MLE loss and perform an SGD update $\theta_{n}^\prime = \theta_n - \alpha \nabla_\theta \mathcal{L}_{\mathrm{MLE}}(\theta_n)$, where $\alpha$ is the step size.
Next, we use the unlikelihood loss on meta-test $(\tilde{x}, y)$ as the loss for the meta optimization w.r.t. the \textit{updated} parameters $\theta^{\prime}_n$:
$
    \mathcal{L}_{\mathrm{UL}}(\theta^{\prime}_n) = - \log (1-p_{\theta_{n}^\prime}(y|\tilde{x}))
$.
Finally, we optimize the sum of the MLE loss $\mathcal{L}_{\mathrm{MLE}}(\theta_n)$ and the meta unlikelihood loss $\mathcal{L}_{\mathrm{UL}}(\theta_{n}^\prime)$ w.r.t. the \textit{original} parameter $\theta_n$ to get the updated parameter $\theta_{n+1}$:
\vspace{-5pt}
\begin{equation*}
\label{eq:meta-ul}
\vspace{-5pt}
  \begin{split}
    \mathcal{L}_{\tau} (\theta_n) &= \mathcal{L}_{\mathrm{MLE}}(\theta_n) + \mathcal{L}_{\mathrm{UL}}(\theta^{\prime}_n) \\
    &= \mathcal{L}_{\mathrm{MLE}}(\theta_n) + \mathcal{L}_{\mathrm{UL}}(\theta_n - \alpha\nabla_\theta \mathcal{L}_{\mathrm{MLE}}(\theta_n)) \\
    \theta_{n+1} &= \theta_n - \beta\nabla_{\theta_n} \mathcal{L}_{\mathrm{\tau}}(\theta_n) 
\vspace{-5pt}
  \end{split}
 \vspace{-5pt}
\end{equation*}
$\alpha$ and $\beta$ are the step sizes for the two updates. A visualization of \abbrmet and \abbrmet-Meta is in~\figref{fig:meta-unlike}.

\paragraph{Selecting the Best Negative Examples $\tilde{x}$.}
Similar to other training schema using multiple examples (e.g., contrastive learning~\cite{gao-etal-2021-simcse}, meta learning~\cite{conklin-etal-2021-meta}, etc.), \abbrmet also depends on the quality of the negative examples $\tilde{x}$. 
One notable difference is that \abbrmet only uses input side $\tilde{x}$, and does not need the corresponding output $\tilde{y}$. This allows us to remove the constraint of only using seen training examples, and to use any good negative examples as long as it is similar to the positive example but semantically different. 
To achieve a similar goal, \citet{conklin-etal-2021-meta} mine examples from the training set via similarity metrics.
However, the input distribution of these mined examples is constrained to the training set, and inherits the highly skewed training distribution.
To overcome this limitation, we directly perturb the original training examples $x$ to create $\tilde{x}$. 
Specifically, we randomly select one token in the training example, and replace the selected word with another word. 
In our preliminary experiments, we notice that it is important (1) to maintain the grammaticality of the perturbed sentence, as training on ungrammatical sentences can make the model easily overfit to the grammar errors; and (2) to have access to the whole vocabulary during training so that we can minimize the influence of a highly skewed training set distribution (a related discussion is at Sec.~\ref{subsec:limit}). 
Hence, we replace the selected words with other words sharing the same grammar properties. 
In this work, we do not delve in the direction of clustering the words with the same grammar properties and use simple approaches to test our ideas. We use an off-the-shelf tagger~\cite{bird2009natural} to obtain the tags of words in COGS, and we cluster the words in SCAN together if they share the same immediate contexts. 
A similar effect can be obtained on more complicated datasets by several different approaches~\cite{stratos2019mutual,akyurek2022compositionality}.

\subsection{Diversifying Lexical Arguments with Primitive Augmentation}
\label{ssec:prim-aug}
While our empirical results demonstrate the effectiveness of \abbrmet, we also notice that \abbrmet does not work on extremely weak baselines. 
In the training set of SCAN \addjump, where a Transformer scores a poor 3.5\% accuracy and does not improve with \abbrmet, each syntactic function (e.g., ``\textit{around left}'') is only paired with four different lexical arguments (``\textit{walk, run, look, turn}'').
Such lack of argument variability in the dataset makes it very easy for the model to memorize all input-output mappings independently.
We hypothesize that current datasets do not provide enough incentives for the model to infer the functional structures with \textit{inductive learning}.
Specifically, in order for a neural model to correctly learn a syntactic function $f$ that can generalize to all the suitable arguments, the dataset must meet two hypotheses.

\begin{condition}[\textbf{Sufficient stimuli for function generalizability}]
$f$ must be applied to a sufficient number of different arguments in the training set.
\label{cond:1}
\end{condition}

Satisfying \condref{cond:1} ensures that models recognize $f$ as a generalizable function instead of a special phrase (e.g., idioms) to memorize independently.
To further generalize $f$ to an unseen argument $a$, the dataset must suffice another hypothesis:

\vspace{-1pt}
\begin{condition}[\textbf{Sufficient stimuli for argument equivalence}]
Suppose during training, function $f$ is applied to a set of arguments $B$. Then in order to apply $f$ to an unseen argument $a$, one of the following two conditions must hold:\label{cond:2}
\end{condition}
\vspace{-5pt}
\begin{Conditions2}
\vspace{-5pt}
\setlength\itemsep{-0.1em}
\item \label{cond:2.1}  \textit{$a$ and another argument $b\in B$ both appear in a sufficient number of distinct functions;}
\item \label{cond:2.2}  \textit{$a$ must appear in the same function with a sufficiently large subset $\bar{B} \subseteq B$.}
\end{Conditions2}
\vspace{-3pt}
This hypothesis makes sure that there is enough stimuli to infer the syntactic equivalence between the new argument $a$ and arguments in $B$, so that $f$ can generalize to $a$.
For example, in SCAN \aroundright, primitives ``\textit{left}'' and ``\textit{right}'' both appear in a number of functions like ``\textit{walk left/right}'', ``\textit{jump opposite left/right}'', which meet \condref{cond:2.1} above and help demonstrate the equivalence of ``\textit{left}'' and ``\textit{right}''.
Therefore, to verify the effectiveness of these hypotheses, we propose the \textbf{prim2primX} data augmentation procedure to automatically enlarge the set of distinct lexical arguments that are applied to a syntactic function.
and lead to empirical improvements in Sec.~\ref{sec:exp}.
We explain this procedure in the next few paragraphs (also see \algref{alg:data-aug} in 
\appendixref{appendix_ssec:methods}).

\paragraph{Building a Lexicon.} 
We first use a dataset-agnostic, rule-based word alignment algorithm to build a lexicon that maps a primitive's input form to its logical form (e.g., ``\textit{run} $\mapsto$ \texttt{RUN}''), while ignoring functional words (e.g., ``\textit{around}'') that only decide the syntactic structure of the outputs.
We denote the source vocabulary as $V$ and the target vocabulary as $W$.
The general intuition is that we want to find pairs $(v,w), v\in V, w\in W$ such that the presence of $v$ in the input is both \textit{necessary} and \textit{sufficient} for the presence of $w$ in the output.
\vspace{-5pt}
\begin{equation} \label{eq:suff-ness}
\vspace{-5pt}
\begin{split}
\mathrm{suff}(v,w) &= \forall (x, y), (v \in x) \xrightarrow{} (w \in y) \\
\mathrm{ness}(v,w) &= \forall (x, y), (w \in y) \xrightarrow{} (v \in x) 
\vspace{-5pt}
\end{split}
\vspace{-5pt}
\end{equation}
Enforcing these two conditions on the SCAN \addjump task returns 6 primitive pairs, e.g., (\textit{run}, \texttt{RUN}), (\textit{walk}, \texttt{WALK}), etc.
Successfully identifying them allows us to later swap them with new primitives to enlarge the set of distinct arguments applied to a function.
However, for tasks with a larger vocabulary, a word's different forms are often parsed to the same output token.
Hence we use the ``\textit{no-winner}'' condition~\cite{akyurek-andreas-2021-lexicon} that only enforces $\mathrm{suff}(v,w)$ and allows many-to-one mappings.
This relaxation makes it possible to build a comprehensive lexicon on more complicated datasets (e.g., COGS) as it allows mappings like ``walk $\mapsto \texttt{WALK}$; walked $\mapsto \texttt{WALK}$'' that share the same target token.
We refer to~\appendixref{appsec:prim-aug} for details.

\paragraph{Mutating Primitives.} 
With a primitive lexicon, we go over each example in the training set and randomly mutate some primitives (prim) by adding a suffix to their source and target forms (primX).
Given an original example ``\textit{walk left twice}'', we select a primitive ``\textit{walk}'' and mutate it to get new examples like ``\textit{\textbf{walk0} left twice} $\mapsto$ \texttt{TL \textbf{WALK0} TL \textbf{WALK0}}''.
This mutation operation can enhance \condref{cond:1} by creating more primitive arguments for each function.
It also improves \condref{cond:2.2} by creating a larger argument set $B$ that is applied to the identity function.
We argue that as the number of distinct arguments for each syntactic function increases, it becomes more challenging to memorize the corresponding output of each input, while the difficulty of compositionally separating function structures from argument symbols remains the same.
Therefore, compositional generalization would become a more favorable solution to the model.
Compared to previous data augmentations~\cite{andreas-2020-good,akyurek2021learning} 
prim2primX has two advantages:
(1) it avoids the difficult task of mining semantically equivalent input-output pairs that can be swapped as it brings in new primitives;
(2) hence the augmented data will never reveal any test compositions, which leads to the model's generalization to future primitives.

%% file: main_tex/4experiments.tex
\begin{table*}[t!]
\centering
\begin{small}
\resizebox{0.93\textwidth}{!}{
\begin{tabular}[t]{l|cccccc}
\toprule
 \centering \textbf{Model} & Overall & C1 & C2 & C3 & C4 & C5 \\
\midrule
MAML~\tiny{\cite{conklin-etal-2021-meta}} & 64.1\tiny{$\pm$3.2} & - & - & - & - & - \\
Lex Learn$^{\ast}$~\tiny{\cite{akyurek-andreas-2021-lexicon}} & 82.0\tiny{$\pm$0.0} & - & - & - & - & - \\
\midrule
\midrule
 Transformer Baseline          &  76.14\tiny{$\pm$2.33} & 82.75\tiny{$\pm$2.25}  &0.00\tiny{$\pm$0.00}  &0.09\tiny{$\pm$0.18}  &93.79\tiny{$\pm$3.57}  &97.39\tiny{$\pm$3.96} \\
\ + MAML                       &  78.33\tiny{$\pm$1.38} & 86.27\tiny{$\pm$1.93}  &0.00\tiny{$\pm$0.00}  &0.12\tiny{$\pm$0.09} & 95.30\tiny{$\pm$2.11}  &99.11\tiny{$\pm$0.48} \\
\ + \abbrmet                     &  \textbf{80.64}\tiny{$\pm$1.08} & \textbf{90.06}\tiny{$\pm$2.09}  &0.00\tiny{$\pm$0.00}  &0.23\tiny{$\pm$0.46}  &\textbf{97.36}\tiny{$\pm$0.97}  & \textbf{99.68}\tiny{$\pm$0.36} \\
\ + \abbrmet-Meta                &  78.14\tiny{$\pm$1.76} & 86.76\tiny{$\pm$3.51}  & 0.00\tiny{$\pm$0.00}  &\textbf{0.73}\tiny{$\pm$0.65}  &93.95\tiny{$\pm$1.68}  &98.64\tiny{$\pm$0.67} \\
 \midrule
 Transformer + prim2primX          &  80.07\tiny{$\pm$1.22} & 88.39\tiny{$\pm$2.48}  & 0.00\tiny{$\pm$0.00}  &0.77\tiny{$\pm$0.64}  &97.59\tiny{$\pm$0.78} & \textbf{99.92}\tiny{$\pm$0.08} \\
 \ + MAML          &  80.93\tiny{$\pm$0.38} & 89.86\tiny{$\pm$0.70}  & 0.00\tiny{$\pm$0.00}  & \textbf{1.19}\tiny{$\pm$1.31}  & \textbf{98.61}\tiny{$\pm$0.35} & 99.20\tiny{$\pm$1.05} \\
\ + \abbrmet                     &  \textbf{81.12}\tiny{$\pm$0.19} & \textbf{90.25}\tiny{$\pm$0.59}  &0.00\tiny{$\pm$0.00}  &0.63\tiny{$\pm$0.58}  &98.55\tiny{$\pm$0.44}  & 99.89\tiny{$\pm$0.12} \\
\ + \abbrmet-Meta                & 80.96\tiny{$\pm$0.43}  & 90.15\tiny{$\pm$0.88}  &0.00\tiny{$\pm$0.00}  &0.52\tiny{$\pm$0.43}  &98.24\tiny{$\pm$0.62}  &99.75\tiny{$\pm$0.44} \\
\bottomrule
\end{tabular}
}
\vspace{-5pt}
\caption{Test accuracy from the COGS \cite{kim-linzen-2020-cogs}, including the overall result on the entire generalization test set and scores on each of the five challenges (C1, C2, C3, C4, C5 as discussed in~\secref{ssec:comp_gen_challenge}). The model with $^{\ast}$ requires special architecture changes, hence is not directly comparable to our numbers. ``+MAML'' is our reimplementation of \citet{conklin-etal-2021-meta} on our own baselines.
}
\label{table:cogs_results}
\vspace{-7pt}
\end{small}
\end{table*}

\section{Experiments}
\label{sec:exp}

\subsection{Experimental Setup and Baselines}

We report results for both LSTMs and Transformers in our SCAN experiments.
We split the original test set into a new development set that contains 10\% of the examples and a new test set containing the rest 90\%.\footnote{In all experiments, we report the mean accuracy in 5 runs and the standard deviation.}
We select the checkpoint with the best dev-set accuracy.
We refer to \appendixref{appendix_ssec:exp_setup} for more details of our experimental setup.

Additionally, in our experiment tables, we provide results for previous representative methods for a more complete picture of the comparison. 
We show results of the meta-learning method used in \citet{conklin-etal-2021-meta} as ``+MAML'' in the tables. 
To the best of our knowledge, this is one of the few baselines that do not require task-specific architecture changes to the model, similar to our method.
Note that while both our method and \citet{conklin-etal-2021-meta} use the MAML framework~\cite{finn2017model}, the training objectives are different: 
the meta-loss in \citet{conklin-etal-2021-meta} is the standard MLE loss, while ours is the unlikelihood loss. We discuss the combination of both ideas in Sec~\ref{subsec:abl}.
Similar to the finding in~\citet{csordas-etal-2021-devil}, we notice different baseline configurations can have a substantial influence on results, hence we report the reimplemented MAML on our baselines using the Levenshtein distance for a fair comparison.\footnote{Additionally, for SCAN MCD and COGS, we also provide the original results from \citet{conklin-etal-2021-meta}. However, \citet{conklin-etal-2021-meta} do not report results on the SCAN \textit{Jump} and \textit{Around Right} splits, so we only compare to our reimplemented results on those two tasks.}
Additionally, for experiments on SCAN, we provide results from previous work with top task-specialized model architecture~\cite{li-etal-2019-compositional,liu2020compositional}, large pretrained model~\cite{furrer2020compositional}, and state-of-the-art data augmentation~\cite{andreas-2020-good}.
On COGS, we provide results for Lex Learn~\cite{akyurek-andreas-2021-lexicon}, which enhances a normal LSTM model with a special copying mechanism.
Many of these baselines (e.g., CGPS, T5-11B, LANE, Lex Learn) require special architecture changes or pretraining models, so they are not directly comparable to our method.

\subsection{COGS Experiment} 
\label{ssec:cogs_results}
Our results on COGS are shown in \tabref{table:cogs_results}, and they also demonstrate the effectiveness of both \abbrmet and prim2primX. 
\abbrmet improves the Transformer performance from 76.14\% to 80.64\%, outperforming MAML and \abbrmet-Meta;
prim2primX can bring additional improvements by creating extra lexical arguments, and pushes the best performance to 81.12\%.
More result discussions and an example showing how prim2primX works on COGS are shown in Appendix~\ref{appsec:cogs}.

Next, to have a better understanding of the behavior of current models, we break down the model accuracy on COGS to five different challenges.
As explained in~\secref{ssec:comp_gen_challenge}, challenges 1, 4, and 5 focus on generalizing a learned syntactic function to an unseen lexical argument, while challenges 2 and 3 focus on generalizing to unseen phrases (structural arguments).
We notice both \abbrmet and prim2primX improve the generalization accuracy on challenges 1, 4, 5, reaching over 90\% on all three challenges, showing promising progress in lexical generalization. 
However, challenges 2 and 3 remain to be two formidable tasks, as all our models 
constantly fall below 1\% accuracy in test examples with novel structural arguments or deeper function recursion.
We discuss this limitation and future remedial approaches in Sec.~\ref{subsec:limit}.

\begin{table}[t!]
\centering
\begin{small}
\resizebox{0.48\textwidth}{!}{%
\begin{tabular}[t]{l|ccc}
\toprule
 \centering \textbf{Model} & \addjump & \aroundright  \\
\midrule
 CGPS$^{\ast}$ \tiny{\cite{li-etal-2019-compositional}} & 98.8\tiny{$\pm$1.4} & 83.2\tiny{$\pm$13.2}\\
 LANE$^{\ast}$~\tiny{\cite{liu2020compositional}} & 100.0 & 100.0 \\
 T5-11B$^{\dagger}$~\tiny{\cite{furrer2020compositional}} & 98.3 & 49.2 \\
 GECA~\tiny{\cite{andreas-2020-good}} & 87\tiny{$\pm$2} & 82\tiny{$\pm$4}\\
 \midrule
\midrule
 LSTM Baseline                & 0.58\tiny{$\pm$0.29} & 10.33\tiny{$\pm$3.20}  \\
 \ + MAML                     & \textbf{0.81}\tiny{$\pm$0.11} & 13.18\tiny{$\pm$4.11} \\
 \ + \abbrmet                 & 0.14\tiny{$\pm$0.07} & \textbf{26.77}\tiny{$\pm$11.14} \\
 \ + \abbrmet-Meta              & 0.21\tiny{$\pm$0.23} & 21.29\tiny{$\pm$7.35}      \\
 \midrule
 Transformer Baseline         & \textbf{3.49}\tiny{$\pm$1.65} & 19.86\tiny{$\pm$10.41}  \\
 \ + MAML                     & 1.87\tiny{$\pm$0.48} & \textbf{41.21}\tiny{$\pm$8.63}   \\
 \ + \abbrmet                 & 1.85\tiny{$\pm$1.79} & 32.41\tiny{$\pm$17.82} \\
 \ + \abbrmet-Meta              & 1.78\tiny{$\pm$1.52} & 36.61\tiny{$\pm$7.16}  \\
 \midrule
 \midrule
 \ LSTM + prim2primX          & 3.34\tiny{$\pm$1.90} & 97.67\tiny{$\pm$1.21}  \\
 \ + MAML                     & 4.51\tiny{$\pm$1.98} & \textbf{99.79}\tiny{$\pm$0.13} \\
 \ + \abbrmet                   & 7.34\tiny{$\pm$5.58} & 97.63\tiny{$\pm$0.99}  \\
 \ + \abbrmet-Meta              & \textbf{7.51}\tiny{$\pm$10.62} & 94.09\tiny{$\pm$5.13}              \\
 \midrule
 \ Transformer + prim2primX     & 63.65\tiny{$\pm$15.47} & 84.50\tiny{$\pm$15.32} \\
  \ + MAML                     & \textbf{76.40}\tiny{$\pm$19.82} & \textbf{99.95}\tiny{$\pm$0.04}\\
 \ + \abbrmet                   & 62.48\tiny{$\pm$9.14} & 62.44\tiny{$\pm$24.36} \\
 \ + \abbrmet-Meta              & 73.94\tiny{$\pm$20.68} & 65.63\tiny{$\pm$20.34}  \\
\bottomrule
\end{tabular}
}
\vspace{-5pt}
\caption{Test accuracy from the SCAN \addjump and \aroundright tasks.  Methods with $^{\ast}$ require special architecture changes and the model with $^{\dagger}$ is pre-trained on large corpora. 
Hence they are not directly comparable to our models.
}
\label{table:scan_results}
\vspace{-7pt}
\end{small}
\end{table}

\begin{table}[t!]
\centering
\begin{small}
\begin{tabular}[t]{l|cc}
\toprule
 \centering \textbf{Model} & \addjump & \adddax \\
\midrule
 Transformer       & 3.49\tiny{$\pm$1.65}  &  0.64 \tiny{$\pm$0.68} \\
 \ + GECA~\tiny{\cite{andreas-2020-good}} & 87\tiny{$\pm$2} & 6.31\tiny{$\pm$7.06} \\
 \ + Recombine~\tiny{\cite{akyurek2021learning}} & \textbf{88.0}\tiny{$\pm$7} &  7.85\tiny{$\pm$3.17} \\
 \ + prim2primX & 63.65\tiny{$\pm$15.47} &  65.35\tiny{$\pm$16.10} \\
 \ + prim2primX + \abbrmet   & 62.48\tiny{$\pm$9.14}   &  \textbf{75.80}\tiny{$\pm$13.82} \\
\bottomrule
\end{tabular}
\vspace{-5pt}
\caption{Test accuracy on generalizing to an unseen primitive ``\textit{dax}''.}
\label{table:dax_results}
\vspace{-7pt}
\end{small}
\end{table}

\subsection{SCAN Results}
\label{ssec:scan_results}

\paragraph{Generalizing functional compositions to new lexical arguments.} 
We report results on SCAN \addjump and \aroundright in \tabref{table:scan_results}.
Both the LSTM and Transformer baselines perform poorly on generalizing learned functions to unseen lexical arguments, reaching < 20\% accuracy on \aroundright and < 4\% accuracy on \addjump.
We first focus on the 2nd and 3rd row-groups in Table~\ref{table:scan_results}. 
On the \aroundright task, with moderate baselines, \abbrmet can improve both Transformer (from 19.86\% to 32.41\%) and LSTM baselines (from 10.33\% to 26.77\%). 
Compared to the other deductive MAML methods~\cite{conklin-etal-2021-meta}, the \abbrmet is better for LSTMs, but worse on Transformers. 
However, with an extremely weak baseline (< 4\% accuracy), none of the deductive methods successfully improves the model on SCAN \addjump.
Next, the bottom two row-groups present the results using prim2primX data augmentation. 
By adding in total 15 additional new lexical arguments (not from the test dataset), 
prim2primX substantially improves the results on both LSTM and Transformer baselines. 
Most notably, it improves the Transformer performance from 3.49\% to 63.65\% on SCAN \addjump, and from 19.86 to 84.50 on \aroundright. 
With stronger baselines on \addjump, our deductive methods can bring additional improvement. 
\abbrmet-Meta improves the performance from 3.34\% to 7.51\% for LSTM, and from 63.65\% to 73.94\% for Transformers. 
On these two tasks, we also see \abbrmet-Meta often outperforms the standard \abbrmet.

\paragraph{Generalizing to future unseen primitives.}
Compared to previous data augmentation methods~\cite{andreas-2020-good,akyurek2021learning},
our prim2primX has one crucial advantage:
as it replaces original words with new primitives, it only encourages the model to induce compositionality from data and does not leak any test compositions.
Oppositely, previous methods learn rules to uncover possible test examples, while models still struggle to generalize beyond the augmented data.
As a result, models trained with prim2primX will generalize to future unseen primitives better.
To demonstrate this, we design a new task \adddax where we add the same number of ``\textit{dax}$\mapsto$\texttt{DAX}'' into a few augmented SCAN \addjump training sets and evaluate on test examples like ``\textit{dax around left twice}''.
\tabref{table:dax_results} show that prim2primX maintains similar improvements
\footnote{For prim2primX, ``\textit{jump}'' and ``\textit{dax}'' are very similar as both primitives exist as stand-alone examples during training. However, due to the high performance variance across different runs on SCAN (also common in other works), \abbrmet shows different performance on \adddax (76.8\%) and \addjump (62.48\%). Nonetheless, the difference here does not influence our claims.} on \adddax as \addjump, while models trained with GECA and Recombine perform poorly on \adddax.

\begin{table}[t!]
\centering
\begin{small}
\resizebox{0.48\textwidth}{!}{%
\begin{tabular}[t]{l|ccc}
\toprule
 \centering \textbf{Model} & MCD1 & MCD2 & MCD3 \\
\midrule
 T5-11B$^{\dagger}$~\tiny{\cite{furrer2020compositional}} & 7.9 & 2.4 & 16.8 \\
 LANE$^{\ast}$~\tiny{\cite{liu2020compositional}} & 100.0 & 100.0 & 100.0 \\
 AuxSeq$^{\ast}$~\tiny{\cite{jiang-bansal-2021-inducing}} & 99.9\tiny{$\pm$0.2} & 90.1\tiny{$\pm$6.5} & 98.2\tiny{$\pm$3.2} \\
 MAML~\tiny{\cite{conklin-etal-2021-meta}} & 47.6\tiny{$\pm$2.3} & 35.2\tiny{$\pm$3.9} & 11.4\tiny{$\pm$3.0} \\
\midrule
\midrule
 LSTM Baseline             & 21.49\tiny{$\pm$3.30} & 13.52\tiny{$\pm$3.49} & 13.11\tiny{$\pm$2.13} \\
\ + MAML                   & 24.27\tiny{$\pm$3.30} & 11.80\tiny{$\pm$2.13} & 12.33\tiny{$\pm$1.37} \\
\ + \abbrmet                 & \textbf{33.20\tiny{$\pm$1.20}} & \textbf{38.66}\tiny{$\pm$2.29} & 14.77\tiny{$\pm$1.84} \\
\ + \abbrmet-Meta            & 31.31\tiny{$\pm$2.23} & 36.20\tiny{$\pm$2.75} & \textbf{15.58}\tiny{$\pm$2.30} \\
\bottomrule
\end{tabular}
}
\vspace{-5pt}
\caption{Test accuracy from the SCAN MCD tasks.}
\label{table:mcd_results}
\vspace{-10pt}
\end{small}
\end{table}

\paragraph{Generalizing to new structural arguments (MCD Tasks).}
The results on three SCAN MCD tasks are presented in~\tabref{table:mcd_results}.
As opposed to the two tasks above that focus on generalizing functions over lexical arguments, the MCD challenges additionally require the model to generalize syntactic functions to \textit{structural} arguments, which contain their own hierarchical structures.
We use LSTM as our baseline because previous work~\cite{conklin-etal-2021-meta} has shown that it outperforms Transformer significantly on SCAN MCD tasks, hence is more suitable for our deductive learning improvements.
On all three tasks, \abbrmet brings substantial improvements to the baseline, with over 12\% of average improvement, and both \abbrmet variants show similar performances and outperform the MAML method~\cite{conklin-etal-2021-meta}.
These results suggest that \abbrmet can improve the generalization to not only new lexical arguments, but also new structural arguments.
We do not observe any improvement by training the LSTM on our augmented data that only create new lexical arguments.
We discuss this limitation and possible future direction in Sec.~\ref{subsec:limit}.

\subsection{Ablation Studies}
\label{subsec:abl}

We investigate the combination of deductive methods (\abbrmet and MAML), and the effect of different numbers of new arguments. More ablations about the unlikelihood loss are in Appendix~\ref{appsec:ul}.

\begin{table}[t!]
\centering
\begin{small}
\resizebox{0.46\textwidth}{!}{
\begin{tabular}[t]{l|cc}
\toprule
 \centering \textbf{Model} & \aroundright (L) & COGS (T) \\
\midrule
\abbrmet               & 25.33\tiny{$\pm$11.16}   &\textbf{ 80.52}\tiny{$\pm$1.13}  \\
MAML                 & 11.77\tiny{$\pm$4.87}    & 78.13\tiny{$\pm$1.33} \\
\abbrmet + MAML        & \textbf{35.85}\tiny{$\pm$19.37}   & 80.37\tiny{$\pm$0.95}  \\
\bottomrule
\end{tabular}
}
\vspace{-5pt}
\caption{Dev set performance change by combining unlikelihood with MAML. L=LSTM, T=Transformer. 
}
\label{table:plus_maml}
\vspace{-6pt}
\end{small}
\end{table}

\begin{table}[t!]
\centering
\begin{small}
\resizebox{0.46\textwidth}{!}{%
\begin{tabular}[t]{l|ccc}
\toprule
 \centering \textbf{Model} & \addjump (T) & \addjump (L) & COGS (T) \\
\midrule
 baseline    & 3.19\tiny{$\pm$1.67}   & 0.70\tiny{$\pm$0.28} & 75.90\tiny{$\pm$2.34} \\
 \ + 1 prim  & 21.66\tiny{$\pm$15.32}  & 0.70\tiny{$\pm$0.23} & 78.82\tiny{$\pm$1.01} \\
 \ + 2 prim  & 77.23\tiny{$\pm$16.64}  & 0.86\tiny{$\pm$0.27} & \textbf{80.14}\tiny{$\pm$1.10} \\
 \ + 3 prim  & \textbf{77.58}\tiny{$\pm$15.78}  & 0.81\tiny{$\pm$0.25} & 78.73\tiny{$\pm$2.36} \\
 \ + 4 prim  & 55.92\tiny{$\pm$8.22}  & 1.17\tiny{$\pm$0.30} & 79.80\tiny{$\pm$1.15} \\
 \ + 5 prim  & 62.80\tiny{$\pm$15.61}  & \textbf{3.30}\tiny{$\pm$1.99} & 79.48\tiny{$\pm$2.08} \\
\bottomrule
\end{tabular}
}
\vspace{-5pt}
\caption{prim2primX ablation, reported as dev accuracy from SCAN \addjump and COGS. `+ 2 prim' means 2 new primitives per original primitive (e.g., run0 and run1).
}
\label{table:data_aug_ablations}
\vspace{-10pt}
\end{small}
\end{table}

\paragraph{Combination of \abbrmet and MAML.}
Both \abbrmet and the MAML loss used in \citet{conklin-etal-2021-meta} aim to improve the deduction learning ability of models. 
Since the two methods come from two different angles, in theory they should also be complementary to each other. 
However, in~\tabref{table:plus_maml}, we see a mixed trend by combining \abbrmet and MAML. 
For example, on COGS, using both unlikelihood and MAML provides no additional gain, while there does appear to be sizable improvements on SCAN \aroundright, with around 10\% improvement on \abbrmet and 24\% over MAML.
These mixed trends indicate that while two deductive methods can have complementary improvements in some situations, 
but there also exist common limitation that is unsolvable to both methods and can lead to a diminishing gain from combining them. 

\paragraph{Data Augmentation Ablation.}
We study how the different number of new arguments in prim2primX improves \textit{inductive learning} of a Transformer.
In~\tabref{table:data_aug_ablations}, there is an overall trend that the model's performance improves as the number of new primitives and total training data increases. However, the gain from using additional data also saturates after a certain amount.
The Transformer baseline requires 2 new primitives per original one to converge to its best performance on SCAN \addjump and COGS, while LSTM needs 5 new primitives to achieve a significant improvement on \addjump. Hence, in our experiments, we use 2 new primitives for COGS, and 5 new primitives for SCAN, where LSTM reaches its best performance, and Transformers still have decent accuracy.

\begin{table}[t!]
\centering
\begin{small}
\resizebox{0.47\textwidth}{!}{%
\begin{tabular}[t]{l|cc}
\toprule
 \centering \textbf{Model} & Structural Error & Primitive Error \\
\midrule
Transformer   & 73.80\tiny{$\pm$21.21}  &  22.99\tiny{$\pm$19.66} \\
 \ + prim2primX  & 35.82\tiny{$\pm$17.36}  &  0.52\tiny{$\pm$0.67} \\
 \ + prim2primX + \abbrmet-Meta & 25.90\tiny{$\pm$23.03} &  0.16\tiny{$\pm$0.21} \\
\bottomrule
\end{tabular}
}
\vspace{-5pt}
\caption{Structural and primitive error of best Transformer models on SCAN 
\addjump.
}
\label{table:error_rate}
\vspace{-9pt}
\end{small}
\end{table}

\subsection{Qualitative Analysis}
Finally, we present a qualitative analysis into how effective are \abbrmet and prim2primX in inducing compositionality in Transformer and reduce the errors discussed in~\secref{sec:error_ana}.
Specifically, we inspect the generated outputs of the multiple best-seed models on SCAN \addjump test set in~\tabref{table:error_rate}. 
We categorize model errors into two categories: primitive errors, and structural errors.
In primitive errors, the models generate the correct structure, but only choose the wrong primitive (e.g., the model generates the output of ``\textit{walks left twice}'' when the input is ``\textit{jump left twice}'').
In structural errors, the models make more significant errors in the generated structure of the output.
In \tabref{table:error_rate}, 22.99\% of the Transformer baseline's errors are primitive errors.
However, Training the model with prim2primX can reduce the error rate to 0.52\%, and further to 0.16\% by also using \abbrmet-Meta.
The improvements suggest that prim2primX and \abbrmet can effectively prevent the model from associating a familiar concept to a new expression.
Our methods reduce the structural error rate to 25.9\% but can not completely eliminate them.
How to further reduce structural errors remains a future direction.

%% file: main_tex/5related.tex
\section{Related Work}

\paragraph{Compositional Generalization.}
Early work has studied neural networks' ability in systematic behavior~\cite{wong2007generalisation,brakel2009strong} in language learning, compositional counting ability~\cite{wiles1998recurrent,weiss-etal-2018-practical} and syntax learning ability~\cite{linzen-etal-2016-assessing}. 
Many recent compositional generalization datasets~\cite{lake2018generalization,kim-linzen-2020-cogs,loula-etal-2018-rearranging,livska2018memorize,bastings-etal-2018-jump,keysers2020measuring,tsarkov2020cfq} greatly facilitate research in this direction.
Previous research explored different ways to tackle the compositionality challenge, including grammar-based approaches~\cite{nye2020learning}, separating syntax and semantics~\cite{li-etal-2019-compositional,russin-etal-2020-compositional}, architecture improvements~\cite{dessi-baroni-2019-cnns,Gordon2020Permutation,oren-etal-2020-improving,zheng2021compositional,herzig2021unlocking,shaw2020compositional}, task decomposition~\cite{liu2020compositional,liu-etal-2021-learning-algebraic}, semi-supervised learning~\cite{guo2021revisiting}, multi-task learning~\cite{jiang-bansal-2021-inducing}, meta-learning~\cite{lake2019metaseq2seq,conklin-etal-2021-meta}, etc.

\vspace{2pt}
\noindent\textbf{Data Augmentation.}
Most data augmentation~\cite{andreas-2020-good,akyurek2021learning} methods promote generalization by creating extra data that resemble the test-set distribution, hence simplifying the problem.
\citet{qiu2021improving} sample new combinations of structures from a context-free grammar induced from the training data.
Contemporarily, \citet{akyurek2022compositionality} adopt a similar procedure to ours. They detect a primitive lexicon and then swap an original primitive with another one of the same types.
In comparison, prim2primX focuses on introducing new primitives, and hence
not exposing any novel test-set compositions.  
Also concurrently\footnote{Our work was first submitted to ACL Rolling Review on March 15, 2022.}, \citet{patel2022revisiting} create new primitives (e.g., ``swim'', ``clap'') and replace the original primitives (e.g., ``walk'') with these new ones to augment the training data.
Similar to our findings, they reveal that the number of different lexical arguments greatly affects the model's generalization.
They further show the limit of models under MLE training on SCAN \textit{Jump} using larger models and much more augmented primitives. 
We instead focus on proving the effectiveness of inductive and deductive methods on widely-used baseline configurations and multiple datasets.

\vspace{2pt}
\noindent\textbf{Mutual Exclusivity Bias.}
Mutual exclusivity (ME) bias helps children acquire the meaning of new words~\cite{markman1988children}. However, vanilla neural networks do not possess this property~\cite{gandhi2020mutual}. Recently, a number of works explore to inject ME bias into the models, mostly focusing on cross-situation word learning~\cite{kadar2015learning, lazaridou2016multimodal, DBLP:conf/cogsci/GulordavaBB20}. 
On a high level, our work is most similar to \citet{DBLP:conf/cogsci/GulordavaBB20}, where they use a margin-based loss for feed-forward networks in a word learning experiment. 
In comparison, our work uses an unlikelihood-based method for seq2seq generation problems. Some compositional generalization methods are also connected to ME biases~\cite{lake2019metaseq2seq,akyurek-andreas-2021-lexicon}, but their specific architecture may not be generalizable to standard seq2seq models.

%% file: main_tex/6conclusion.tex
\section{Conclusion}
We propose two methods to improve compositional generalization: (1) \abbrmet, a mutual exclusivity bias-inspired method implemented with unlikelihood training; and (2) prim2primX, a data augmentation procedure automatically diversifying the arguments of syntactic functions. Empirical results demonstrate the effectiveness of both methods.

%% file: main_tex/7limitations.tex
\section{Limitations}
\label{subsec:limit}
In this section, we discuss the limitations of our methods and point out promising future directions and challenges.
In our experiments, we observe \abbrmet does provide substantial improvements on SCAN and COGS by injecting a certain amount of ME bias into the models. Nonetheless, the effect of ME bias achieved by \abbrmet is not equal to the ME bias of humans. \abbrmet regularizes the model to not generate seen outputs (or expressions) facing unseen inputs (or concepts). Human ME biases, however, will additionally let humans assign two different unseen expressions (or outputs) to two different unseen concepts (or inputs). To summarize, \abbrmet achieves ME bias on the trained domain, which already leads to substantial improvement on SCAN and COGS. Future work should explore directions to achieve generalizable ME bias, which may lead to a further breakthrough in this task.

Additionally, as discussed in~\secref{ssec:cogs_results}, one remaining challenge of our models is to improve generalization to novel structural arguments, where all the models achieve close to zero accuracy. 
Similar to the trends on other datasets, the extremely low performance of the baselines makes it hard for our deductive method \abbrmet to improve.
On the other hand, our prim2primx data augmentation has the potential to improve similar problems, as reflected in the SCAN \addjump results. 
However, the success depends on automatically identifying \textit{lexicon}-level arguments. 
In contrast, automatically identifying any \textit{phrase}-level structures is still an open challenge, and we leave this for future work.

%% file: main_tex/8ethics.tex
\section{Ethical Considerations}
The methods proposed in this paper aim to improve the compositional generalization ability of the models. Progress in this direction can lead to more systematic and predictable behavior of neural models in downstream applications. In this work, improvements are evaluated on two highly-controlled datasets, SCAN~\cite{lake2018generalization} and COGS~\cite{kim-linzen-2020-cogs} as proof-of-concept experiments. We also point out potential challenges and directions to apply our methods in practice in Sec.~\ref{subsec:limit}.

%% file: appendix_tex/1methods.tex
\section{Methods}
\label{appendix_ssec:methods}

\subsection{Diversify Lexical Arguments with Primitive Augmentation}
\label{appsec:prim-aug}
\paragraph{Building a Lexicon.}
Enforcing the two conditions in~\eqnref{eq:suff-ness} on the SCAN \addjump task would return 6 pairs of primitives, e.g., (\textit{run}, \texttt{RUN}), (\textit{walk}, \texttt{WK}), etc.
However, for other tasks with a larger vocabulary, a word's different surface forms are often parsed to the same output meaning.
For example, in COGS, both ``\textit{paint}'' and ``\textit{painted}'' are mapped to ``\texttt{paint}'' in the output form. 
Therefore, we use the ``\textit{no-winner}'' condition~\cite{akyurek-andreas-2021-lexicon} that allows an $M$-to-one mapping if $v$ is not necessary for $w$ as long as $M <$ threshold value $\psi$.
\begin{equation*} \label{eq:no-winner}
\begin{split}
\mathrm{NoWinner}(w) &= \nexists v, \mathrm{suff}(v,w) \land \mathrm{ness}(v,w)\\
\mathrm{IsPrim}(v,w) &= \mathrm{suff}(v,w) \land \\
&(\mathrm{ness}(v,w) \lor \mathrm{NoWinner}(w))
\end{split}
\end{equation*}
Finally, we delete words above a certain frequency and words below another frequency from the lexicon.
This frequency threshold is decided based on dev-set tuning results.

Compared to the previous compositional data augmentation methods~\cite{andreas-2020-good,akyurek2021learning} that substitute words with other semantically equivalent words from the same lexicon, our method has two main advantages:
(1) it circumvents the difficult task of finding semantically equivalent words that share some common environment and creating templates that can be filled with these words.
(2) because we are not replacing the primitive with another word from the existing lexicon (e.g., \textit{jump}), the augmented data will not expose/reveal any novel compositions (``\textit{jump around left}'') in the test set, and thus still preserve the compositional generalization challenge in the original task.
Another side advantage of creating new primitives instead of swapping with one from the original lexicon is that we can train the model to generalize to future new primitives (e.g., dax) simply by adding ``\textit{dax}$\mapsto$\texttt{DAX}'' to the training set without the need to rerun any data-augmentation procedures.

\begin{algorithm}[t]
\caption{prim2primX Data Augmentation}\label{alg:data-aug}
\begin{algorithmic}
\Require Training data $D$
\Require All primitives $P$
\Require Number of new primitives $n$
\Ensure Augmented data $D_a$
\For{$(x,y) \in D$}
\State $D_a = D_a \cup (x,y)$
\For{token $x_t$ in $x$}
\State $x' \gets x$, $y' \gets y$
\State $y_t \gets P[x_t]$
\If{$t \in P$}
    \State Sample $s$ from $\{0, 1, ..., n\}$
    \If{$s$ = 0}
    break
    \EndIf
    \State Replace all $x_t$ in $x'$ with $x_ts$
    \State Replace all $y_t$ in $y'$ with $y_ts$
\EndIf
\EndFor
\State $D_a = D_a \cup (x',y')$
\EndFor
\end{algorithmic}
\end{algorithm}

%% file: appendix_tex/2experiments.tex
\section{Experimental Details}

\begin{table*}[t]
\centering
\begin{scriptsize}
\begin{tabular}[t]{c|c|c|c}
\toprule
 \centering \multirow{3}{*}{\centering \small Function $f$} & \multicolumn{3}{c}{\small Examples} \\
 \cmidrule{2-4}
  & \textbf{Input} & \textbf{Program} & \textbf{Output} \\
\midrule
 $\mathrm{Identity}(x) = x$ & walk & $\mathrm{Identity}(\mathrm{walk})$ & WK \\  
 $\mathrm{Rev}(x_1, x_2) = x_2 + x_1$ & walk left & $\mathrm{Rev} (\mathrm{walk}, \mathrm{left})$ & TL WK \\  
 $\mathrm{Oppo}(x_1, x_2) = x_2 + x_2 + x_1$ & walk \textbf{opposite} left & $\mathrm{Oppo} (\mathrm{walk}, \mathrm{left})$ & TL TL WK \\  
 $\mathrm{Around}(x_1, x_2) = (x_2 + x_1 ) * 4$ & walk \textbf{around} left & $\mathrm{Around} (\mathrm{walk}, \mathrm{left})$ & TL WK TL WK TL WK TL WK  \\ 
\midrule
 \multirow{2}{*}{$\mathrm{Twice}(x) = x * 2$} & walk \textbf{twice} & $\mathrm{Twice}(\mathrm{walk})$ & \scriptsize WK WK \\
 & walk opposite left \textbf{twice} & $\mathrm{Twice}(\mathrm{Oppo}(\mathrm{walk}))$ & TL TL WK TL TL WK\\
\midrule
 \multirow{2}{*}{$\mathrm{Thrice}(x) = x * 3$} & walk \textbf{thrice} & $\mathrm{Thrice}(\mathrm{walk})$ & \scriptsize WK WK WK \\
 & walk opposite left \textbf{thrice} & $\mathrm{Thrice}(\mathrm{Oppo}(\mathrm{walk}))$ & TL TL WK TL TL WK TL TL WK \\
 \midrule 
 $\mathrm{And}(x_1, x_2) = x_1 + x_2$ & walk left \textbf{and} run & $\mathrm{And} (\mathrm{Rev}(\mathrm{walk}, \mathrm{left}), \mathrm{run})$ & TL WK RUN \\  
  $\mathrm{After}(x_1, x_2) = x_2 + x_1$ & walk left \textbf{after} run & $\mathrm{After} (\mathrm{Rev}(\mathrm{walk}, \mathrm{left}), \mathrm{run})$ & RUN TL WK \\  
\bottomrule
\end{tabular}
\end{scriptsize}
\caption{All syntactic functions in SCAN~\cite{lake2018generalization} and examples of applying these functions to primitive and structural arguments. ``$x_1 + x_2$'' means concatenating $x_1$ and $x_2$ at the output side and $x * 2$ means replicating $x$ twice.
}
\label{apptable:scan_functions}
\end{table*}

\subsection{SCAN Dataset}
\label{appendix_ssec:dataset}
The SCAN dataset~\cite{lake2018generalization} consists of natural language commands paired with action sequences.
Each sub-command is made of three types of words: action primitive (``\textit{walk, jump, look, run, turn}''), direction primitive (``\textit{left, right}''), and functional words (``\textit{opposite, around, twice, thrice}''), and can be connected with another sub-command via a function (``\textit{and, after}''). \tabref{apptable:scan_functions} describes all syntactic functions of SCAN.

\vspace{3pt}
\noindent\textbf{Jump} evaluates the model's ability to generalize compositions of syntactic functions across different lexical (primitive) arguments. 
The training set consists of the primitive command ``\textit{jump}'' on its own, all other primitives, and compound commands \textit{without} ``\textit{jump}'' (e.g., ``\textit{walk around left}'');  the test set contains compound commands with ``\textit{jump}'' (e.g., ``\textit{jump around left}'').

\vspace{3pt}
\noindent\textbf{Around Right}~\cite{loula-etal-2018-rearranging} puts all commands containing templates
of the form ``\textit{[primitive] around right}'' in the test set, with the remaining examples in the training set. 

\vspace{3pt}
\noindent\textbf{MCD} splits are created to maximize the output compound divergence while guaranteeing a small atom divergence between train and test sets \cite{keysers2020measuring}.
For example, the training and dev sets of MCD1 have a similar distribution of individual words to ensure minimal atom divergence.
However, the training set does not contain compounds ``\textit{[primitive] around left twice}'', which only appear in the dev and test sets.
These splits require a higher level of compositionality than recognizing the syntactic equivalence of primitives:
the models must be able to 
(1) understand the underlying symbolic functions ``$x$ twice$\mapsto x\;x$'' from training examples like ``\textit{jump left twice}'' and master the semantics of ``\textit{jump around left}'' from examples like ``\textit{jump around left thrice}'';
(2) compositionally apply the function ``twice'' to a novel argument ``\textit{jump around left}'' in the dev and test sets.

\subsection{COGS Dataset} 
The COGS dataset~\cite{kim-linzen-2020-cogs} requires parsing a diverse set of natural language sentences into their corresponding logical forms based on lambda calculus to accurately reflect the semantic representation of the natural sentence. 
COGS raises five different systematic generalization challenges in its test set: 
(1) novel combination of familiar primitives and syntactic functions; 
(2) novel combination of modified phrases and syntactic functions; 
(3) sentences with deeper recursion; 
(4) sentences with alternative verb argument structures (e.g., active-passive), and 
(5) novel identity of a verb (e.g., unaccusative and unergative).
Challenges 1, 4, and 5 require generalizing a syntactic function to a lexical argument which was never associated with this function during training, 
while challenges 2 and 3 require generalizing to unseen structural arguments.

\subsection{Experimental Setup}
\label{appendix_ssec:exp_setup}
\paragraph{Baselines.}
We report results for both LSTMs and Transformers in our experiments. Our LSTM hyper-parameters are adopted from \citet{akyurek-andreas-2021-lexicon}, with 2 layers in both encoder and decoder, a hidden dimension size of 512, and a dropout rate of 0.4.
Our Transformer hyperparameters and configurations are set following the findings in \citet{csordas-etal-2021-devil}. We use relative positional embedding~\cite{shaw-etal-2018-self} and use 3 layers in both encoder and decoder. The hidden dimension size and the embedding size are set to 256, and the dimension of the feed-forward layer is set to 512. We use a dropout rate of 0.1 and 4 self-attention heads. The Transformer model is implemented using OpenNMT~\cite{klein-etal-2017-opennmt}.
Additionally, we also report the performance of the meta-learning method used in \citet{conklin-etal-2021-meta} as our baseline, and will be shown as ``+MAML'' in the result tables. Since in our preliminary experiments, we notice different baselines have a substantial influence on the performance~\cite{csordas-etal-2021-devil}, we report the reimplemented performance of their method using the Levenshtein distance on the previously described LSTM and Transformer baselines for a fair comparison with other approaches. 

\paragraph{Training Details.}
Recent works~\cite{csordas-etal-2021-devil,conklin-etal-2021-meta} observed that the in-domain development set cannot provide enough signal for doing model selection for compositional generalization. Hence, similar to \citet{conklin-etal-2021-meta}, we split the original test set into a new development set that contains 10\% of the examples and a new test set containing the rest 90\%. We select the checkpoint with the best accuracy on the development set. In all our experiments, we report the mean performance in 5 runs and the corresponding standard deviation.

For all our LSTM models, we train our model using a dropout rate of 0.4, a peak learning rate of 1.0 with the Noam learning rate scheduling. The baseline model is trained using a batch size of 512, and for 8000 steps with 4000 warm-up steps on the SCAN datasets. For the primitive augmentation datasets, we notice the model needs a longer time to converge since the dataset is multiple times larger, so we train those models for 40000 steps with 4000 warm-up steps.
For all our Transformer models, we train our model using a dropout rate of 0.1, a peak learning rate of 2.0 with the Noam learning rate scheduling. The model is trained using a batch size of 128, and for 50000 steps with 5000 warm-up steps in all the experiments. All our models can be trained on a single NVIDIA Titan Xp GPU.

%% file: appendix_tex/3ulcomparison.tex
\begin{table}[t!]
\centering
\begin{small}
\begin{tabular}[t]{l|c}
\toprule
 \centering \textbf{Model} & Performance \\
\midrule
 Transformer          &  62.84\tiny{$\pm$9.54} \\
 \ + \abbrmet      &  100.00\tiny{$\pm$0.00} \\
 \ + \abbrmet-Meta &  100.00\tiny{$\pm$0.00} \\
\bottomrule
\end{tabular}
\vspace{-5pt}
\caption{The performance of the Transformer baseline and \abbrmet on the synthetic task described in Sec.~\ref{sec:toy}}.
\label{table:met_toy}
\vspace{-5pt}
\end{small}
\end{table}

\begin{table}[t!]
\centering
\begin{small}
\begin{tabular}[t]{l|cc}
\toprule
 \centering \textbf{Model} & \addjump$^\star$ & COGS \\
\midrule
 Transformer          & 62.80\tiny{$\pm$15.61}  &  75.90\tiny{$\pm$2.34} \\
 \ + Avg-Word Unlike  & 1.32\tiny{$\pm$0.43}   &  6.16\tiny{$\pm$1.39} \\
 \ + Min-Word Unlike  & 59.97\tiny{$\pm$8.60}  &  \textbf{80.58}\tiny{$\pm$0.78} \\
 \ + Sent Unlike      & 60.68\tiny{$\pm$8.96}  &  80.52\tiny{$\pm$1.13} \\
 \ + Meta Sent Unlike & \textbf{73.30}\tiny{$\pm$21.04}  &  79.44\tiny{$\pm$1.39} \\
\bottomrule
\end{tabular}
\vspace{-5pt}
\caption{Unlikelihood loss ablation, reported as dev accuracy on SCAN \addjump and COGS. 
Models for \addjump are trained on augmented data.
}
\label{table:unlike_ablations}
\vspace{-5pt}
\end{small}
\end{table}

\section{Additional Results}
\label{app:met}
\subsection{Encouraging Mutual Exclusivity in a Synthetic Experiment}
\label{sec:toy}
In this experiment, we design a highly-controlled synthetic experiment to demonstrate how \abbrmet encourages mutual exclusivity in the model. Our experiment design is adapted from the toy experiment design in \citet{zheng-lapata-2022-disentangled}. 

We construct the dataset using two bijections: $f_1$ between a source vocabulary $S_1$ and a target vocabulary $T_1$, and $f_2$ between vocabulary $S_2$ and $T_2$. The set $S_1$ can be further split to two disjoint sets $S_1^{train}$ and $S_1^{gen}$, each mapping to $T_1^{train}$ and $T_1^{gen}$ respectively. 

Following \citet{zheng-lapata-2022-disentangled}, we construct our training set so the model can exploit some spurious correlations between the input and the output. Specifically, the training set will contain two types of examples. For the first type of examples, the input only contain one token $s_1 \in S_1$ the output is $f_1(s_1)$. For the second type of examples, the input sequence contains two tokens `$s_1^{train},s_2$', where $s_1^{train}\in S_1^{train}$ and $s_2\in S_2$. The corresponding output is `$f_1(s_1^{train}), f_2(s_2)$'.

All the testing examples will look similar to the second type and contain two tokens, $s_1$ and $s_2$, where $s_1\in S_1$ and $s_2\in S_2$. Crucially, the main difficulty of this testing set lies in predicting the correct outputs for inputs that \textbf{never} appears in the training set, i.e. inputs containing $s_1^{gen}$ and $s_2$, while $s_1^{gen}$ never appears in the second type of examples in the training set. 

The performance of the Transformer baseline is shown in Table~\ref{table:met_toy}. We can see that despite the simplicity of this task, the Transformer baseline struggles to perform well on the novel examples, only reaching 62.80 accuracy. A common failure case for the baselines is to only predict $f_1(s_1^{gen})$ for the input sequence `$s_1^{gen}, s_2$' and completely ignores the second input token. This is because during training, $s_1^{gen}$ only appears in sentences with total length 1, and the model can exploit this spurious pattern. By applying the \abbrmet framework, we can prevent this incorrect behavior, since the output $f_1(s_1^{gen})$ should only be matched with the input $s_1^{gen}$ and should not be paired with any other input. The empirical improvements on this task confirm the effectiveness of \abbrmet. From Table~\ref{table:met_toy}, we can see using both \abbrmet or \abbrmet-Meta can make the model reach 100\% accuracy and completely solves this task.

\subsection{Ablation Experiments on the Implementation of the Unlikelihood Loss}
\label{appsec:ul}
Notably, the unlikelihood loss $\mathcal{L}_{\mathrm{UL}}$ implementation used in \abbrmet is different from the loss used in \citet{welleck2019neural}, as the original unlikelihood is an average over all possible locations, while the unlikelihood loss in Eq.~\ref{eq:ul} is operated at the sentence level.
We make this design choice since our unlikelihood is not used to penalize the generation of each token (which is the target in \citet{welleck2019neural}), but to penalize the generation of the whole sentence, in which only several key tokens are wrong and should be penalized. The sentence-level unlikelihood loss can achieve this by only maximizing the unlikelihood on a few words, while keeping the likelihood on most words unchanged. Our ablation experiments below also verify the advantage of this design choice.

\paragraph{Empirical Comparison of Different Loss Functions}
In Table~\ref{table:unlike_ablations}, we compare the effect of different unlikelihood loss variants.
On both SCAN \addjump and COGS, our sentence-level variant (Sent Unlike) is substantially better than the original variant used in \citet{welleck2019neural} that averages unlikelihood loss at the word level (Avg-Word Unlike), as the penalization effect of the latter method is too strong and lead to a very bad final performance. In our preliminary experiments, we have also tried to use a smaller weight coefficient for the Avg-Word Unlike variant. While using a smaller coefficient is beneficial, the Sent Unlike variant still shows an advantage.
We also experimented with another variant where we still compute the unlikelihood loss at the word level, but use a min-pooling operation to get the final loss, which achieves a similar effect as our Sent Unlike loss since the penalization will only apply to a certain token. From Table~\ref{table:unlike_ablations}, we can see this variant shows similar performance to the Sent Unlike, further verifying our hypothesis about why Sent Unlike is preferable to the original variant in \citet{welleck2019neural} in our experiments.

\section{Examples for the effect of prim2primX on COGS}
\label{appsec:cogs}
Given a training example ``\textit{A rose was helped by a dog}'', our prim2primX data augmentation will first identifies ``\textit{rose}'', ``\textit{helped}'', and ``\textit{dog}'' as primitives,
and then randomly swap some of these primitives with their mutated forms to create new training examples (e.g., ``\textit{A \textbf{rose1} was \textbf{helped0} by a dog.}'').
The quantitative results on COGS are in Table~\ref{table:cogs_results} in the main paper. Just by using the prim2primX data augmentation, we can improve the Transformer baseline from 76.14 \% accuracy to 80.07\% accuracy, showing a 3.93\% absolute improvement from the model trained with original data.
\abbrmet further boosts the accuracy to 81.12\%, again demonstrating that \textit{deductive} \abbrmet method can provide complementary performance gain on top of \textit{inductive} data augmentation.